\documentclass[11pt,a4paper]{article}
\usepackage[hyperref]{emnlp2020}
\usepackage{times}
\usepackage{latexsym}
\usepackage{amssymb}
\usepackage{amsthm}
\usepackage{mathtools}
\usepackage{multirow}
\usepackage{microtype}
\usepackage{amssymb}
\usepackage{graphicx}
\usepackage{bm}
\usepackage{multirow}
\usepackage{pstool}
\usepackage{subfigure}

\newcommand{\real}[1]{\mathbb{R}^{#1}}

\usepackage{microtype}
\usepackage{algorithm}
\usepackage{algpseudocode}
\aclfinalcopy

\title{Tell Me How to Ask Again: Question Data Augmentation \\ with Controllable Rewriting in Continuous Space}

\author{Dayiheng Liu$^\spadesuit$\thanks{\hspace{2mm}Work is done during internship at Microsoft Research Asia.}\:\:\:\: Yeyun Gong$^{\dag}$ \:\:\:\:  Jie Fu$^\diamondsuit$ \:\:\:\: Yu Yan$^{\ddag}$ \\ 
\textbf{Jiusheng Chen$^{\ddag}$ \:\:\:\: Jiancheng Lv$^{\spadesuit}$ \:\:\:\: Nan Duan$^{\dag}$ \:\:\:\: Ming Zhou$^{\dag}$}   \\
$^\spadesuit$College of Computer Science, Sichuan University    \:\:\:\:  $^\dag$Microsoft Research Asia  \\  $^\diamondsuit$ Mila \:\:\:\:  $^\ddag$Microsoft \\
losinuris@gmail.com \:\:\:\: 
}

\date{}

\begin{document}
\maketitle
\begin{abstract}
In this paper, we propose a novel data augmentation method, referred to as Controllable Rewriting based Question Data Augmentation (CRQDA), for machine reading comprehension (MRC), question generation, and question-answering natural language inference tasks. We treat the question data augmentation task as a constrained question rewriting problem to generate context-relevant, high-quality, and diverse question data samples. 
CRQDA utilizes a Transformer autoencoder to map the original discrete question into a continuous embedding space.
It then uses a pre-trained MRC model to revise the question representation iteratively with gradient-based optimization.
Finally, the revised question representations are mapped back into the discrete space, which serve as additional question data.
Comprehensive experiments on SQuAD 2.0, SQuAD 1.1 question generation, and QNLI tasks demonstrate the effectiveness of CRQDA\footnote{The source code and dataset will be available at \url{https://github.com/dayihengliu/CRQDA}.}.
\end{abstract}
 
\section{Introduction}
Data augmentation (DA) is commonly used to improve the generalization ability and robustness of models by generating more training examples.
Compared with the DA used in the fields of computer vision~\cite{krizhevsky2012imagenet,szegedy2015going,cubuk2019autoaugment} and speech processing~\cite{ko2015audio}, how to design effective DA tailored to natural language processing (NLP) tasks remains a challenging problem. 
Unlike the general image DA techniques such as rotation and cropping, it is more difficult to synthesize new high-quality and diverse text.

Recently, some textual DA techniques have been proposed for NLP, which mainly focus on text classification and machine translation tasks. 
One way is directly modifying the text data locally with word deleting, word order changing, and word replacement~\cite{fadaee2017data,kobayashi2018contextual,wei2019eda,wu2019conditional}.
Another popular way is to utilize the generative model to generate new text data, such as back-translation~\cite{sennrich2015improving,yu2018qanet}, data noising technique~\cite{xie2017data}, and utilizing pre-trained language generation model~\cite{kumar2020data,anaby2020not}.

Machine reading comprehension (MRC)~\cite{rajpurkar2018know}, question generation (QG)~\cite{du2017learning, zhao2018paragraph} and, question-answering natural language inference (QNLI)~\cite{demszky2018transforming,wang2018glue} are receiving attention in NLP community. MRC requires the model to find the answer given a paragraph\footnote{It can also be a document span or a passage. For notational simplicity, we use the ``paragraph'' to refer to it in the rest of this paper.} and a question, while QG aims to generate the question for a given paragraph with or without a given answer. Given a question and a sentence in the relevant paragraph, QNLI requires the model to infer whether the sentence contains the answer to the question. 
Because the above tasks require the model to reason about the question-paragraph pair, existing textual DA methods that directly augment question or paragraph data alone may result in irrelevant question-paragraph pairs, which cannot improve the downstream model performance. 

Question data augmentation (QDA) aims to automatically generate context-relevant questions to further improve the model performance for the above tasks~\cite{yang2019xlnet,dong2019unified}. 
Existing QDA methods mainly employ the round-trip consistency~\cite{alberti2019synthetic,dong2019unified} to synthesize answerable questions. However, the round-trip consistency method is not able to generate context-relevant unanswerable questions, where MRC with unanswerable questions is a challenging task~\cite{rajpurkar2018know,kwiatkowski2019natural}. 
\citet{zhu2019learning} firstly study unanswerable question DA, which relies on annotated plausible answer to constructs a small pseudo parallel corpus of answerable-to-unanswerable questions for unanswerable question generation. Unfortunately, most question answering (QA) and MRC datasets do not provide such annotated plausible answers.

Inspired by the recent progress in controllable text revision and text attribute transfer~\cite{wang2019controllable,liu2020revision}, we propose a new QDA method called \textbf{C}ontrollable \textbf{R}ewriting based \textbf{Q}uestion \textbf{D}ata \textbf{A}ugmentation (\textbf{CRQDA}), which can generate both new context-relevant answerable questions and unanswerable questions. The main idea of CRQDA is to treat the QDA task as a constrained question rewriting problem. Instead of revising \textit{discrete} question directly, CRQDA aims to revise the original questions in a \textit{continuous embedding space} under the guidance of a pre-trained MRC model. 
There are two components of CRQDA: (i) A Transformer-based autoencoder whose encoder maps the question into a latent representation. Then its decoder reconstructs the question from the latent representation. (ii) A MRC model, which is pre-trained on the original dataset. This MRC model is used to \textit{tell} us how to revise the question representation so that the reconstructed new question is a context-relevant unanswerable or answerable question.
The original question is first mapped into a continuous embedding space.
Next, the pre-trained MRC model provides the guidance to revise the question representation iteratively with gradient-based optimization.
Finally, the revised question representations are mapped back into the discrete space, which act as the additional question data for training.

In summary, our contributions are as follows: (1) We propose a novel controllable rewriting based QDA method, which can generate additional high-quality, context-relevant, and diverse answerable and unanswerable questions. (2) We compare the proposed CRQDA with state-of-the-art textual DA methods on SQuAD 2.0 dataset, and CRQDA outperforms all those strong baselines consistently. (3) In addition to MRC tasks, we further apply CRQDA to question generation and QNLI tasks, and comprehensive experiments demonstrate its effectiveness.

\section{Related Works}
Recently, textual data augmentation has attracted a lot of attention. One popular class of textual DA methods is confined to locally modifying text in the discrete space to synthesize new data. \citet{wei2019eda} propose a universal DA technique for NLP called easy data augmentation (EDA), which performs synonym replacement, random insertion, random swap, or random deletion operation to modify the original text. \citet{Jungiewicz2019TowardsTD} propose a word synonym replacement method with WordNet. \citet{kobayashi2018contextual} relies on word paradigmatic relations. More recently, CBERT~\cite{wu2019conditional} retrofits BERT~\cite{devlin2018bert} to conditional BERT to predict the masked tokens for word replacement. These DA methods are mainly designed for the text classification tasks.

Unlike modifying a few local words, another commonly used textual DA way is to use a generative model to generate the entire new textual samples, including using variational autoencodes (VAEs)~\cite{kingma2013auto,rezende2014stochastic}, generative adversarial networks (GANs)~\cite{tanaka2019data}, and pre-trained language generation models~\cite{radford2019language,kumar2020data,anaby2020not}. 
Back-translation~\cite{sennrich2015improving,yu2018qanet} is also a major way for textual DA, which uses machine translation model to translate English sentences into another language (e.g., French), and back into English. 
Besides, data noising techniques~\cite{xie2017data,marivate2019improving} and paraphrasing~\cite{kumar2019submodular} are proposed to generate new textual samples. 
All the methods mentioned above usually generate individual sentences separately.
For QDA of MRC, QG, and QNLI tasks, these DA approaches cannot guarantee the generating question are relevant to the given paragraph.
In order to generate context-relevant answerable and unanswerable questions, our CRQDA method utilizes a pre-trained MRC as guidance to revise the question in continuous embedding space, which can be seen as a special \textit{constrained} paraphrasing method for QDA. 

Question generation~\cite{heilman2010good,du2017learning,zhao2018paragraph,zhang2019addressing} is attracting attention in the field of natural language generation (NLG). 
However, most previous works are not designed for QDA. 
That is, they do not aim to generate context-relevant questions for improving downstream model performance. 
Compared to QG, QDA is relatively unexplored. Recently, some works~\cite{alberti2019synthetic,dong2019unified} utilize round-trip consistency technique to synthesize answerable questions. 
They first use a generative model to generate the question with the paragraph and answer as model input, and then use a pre-trained MRC model to filter the synthetic question data.
However, they are unable to generate context-relevant unanswerable questions. It should be noted that our method and round-trip consistency are orthogonal. 
CRQDA can also rewrite the synthetic question data by other methods to obtain new answerable and unanswerable question data.
Unanswerable QDA is firstly explored in~\citet{zhu2019learning}, which constructs a small pseudo parallel corpus of paired answerable and unanswerable questions and then generates relevant unanswerable questions in a supervised manner. This method relies on annotated plausible answers for the unanswerable questions, which does not exist in most QA and MRC datasets. 
Instead, our method rewrites the original answerable question to a relevant unanswerable question in an unsupervised paradigm, which can also rewrite the original answerable question to another new relevant answerable question.

Our method is inspired by the recent progress on controllable text revision and text attribute transfer~\cite{wang2019controllable,liu2020revision}. However, our approach differs in several ways. First, those methods are used to transfer the attribute of the single sentence alone, but our method considers the given paragraph to rewrite the context-relevant question.
Second, existing methods jointly train an attribute classifier to revise the sentence representation, while our method unitizes a pre-trained MRC model that shares the embedding space with autoencoder as the guidance to revise the question representation. Finally, the generated questions by our method serve as augmented data can benefit the downstream tasks.

\begin{figure*}[th]
    \centering
	\includegraphics[width = 6.5 in]{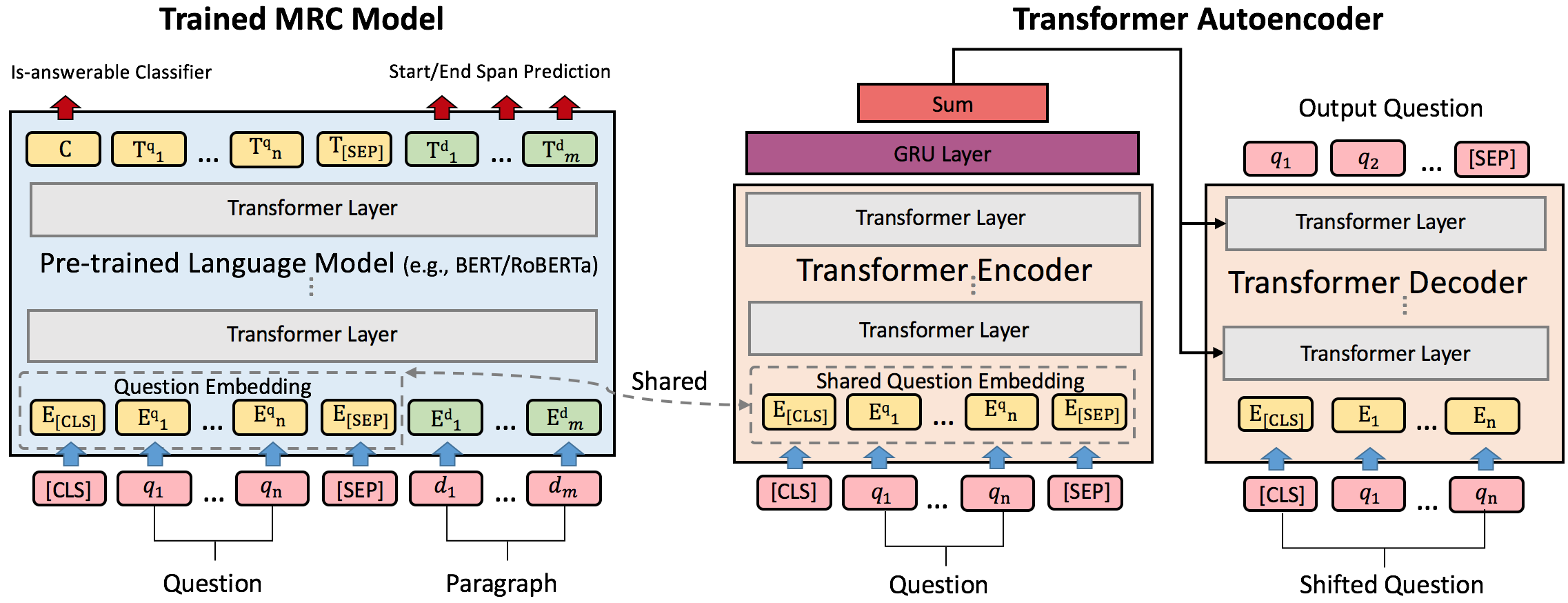}
	\caption{The architecture of CRQDA. }\label{fig:overall}
\end{figure*}

\section{Methodology}
\subsection{Problem Formulation}
We consider an extractive MRC dataset $\mathcal{D}$, such as SQuAD 2.0~\cite{rajpurkar2018know}, which has $|\mathcal{D}|$ 5-tuple data: $(q,d,s,e,t)$, where $|\mathcal{D}|$ is the data size, $q=\{q_1,...,q_n\}$ is a tokenized question with length $n$, $d=\{d_1,...,d_m\}$ is a tokenized paragraph with length $m$, $s,e \in \{0, 1, ..., m-1\}$ are inclusive indices pointing to the start and end of the answer span, and $t \in \{0, 1\} $ represents whether the question $q$ is answerable or unanswerable with $d$.
Given a data tuple $(q,d,s,e,t)$, we aim to rewrite $q$ to a new answerable or unanswerable question $q'$ and obtain a new data tuple $(q',d,s,e,t')$ that fulfills certain requirements: 
(\textit{i}) The generated answerable question can be answered with the answer span $(s,e)$ with $d$, while the generated unanswerable question cannot be answered with $d$. (\textit{ii}) The generated question should be relevant to the original question $q$ and paragraph $d$. (\textit{iii}) The augmented dataset $\mathcal{D}'$ should be able to further improve the performance of the MRC models.

\subsection{Method Overview}
Figure~\ref{fig:overall} shows the overall architecture of CRQDA. The proposed model consists of two components: a pre-trained language model based MRC model as described in \S~\ref{sec:mrc}, and a Transformer-based autoencoder as introduced in \S~\ref{sec:ae}. Given a question $q$ from the original dataset $\mathcal{D}$, we first map the question $q$ into a continuous embedding space. Then we revise the question embeddings by gradient-based optimization with the guidance of the MRC model  (\S~\ref{sec:inf}). Finally, the revised question embeddings are inputted to the Transformer-based autoencoder to generate a new question data.

\subsection{Pre-trained Language Model based MRC Model}\label{sec:mrc}
In this paper, we adopt the pre-trained language model (e.g., BERT~\cite{devlin2018bert}, RoBERTa~\cite{liu2019roberta}) based MRC models as our MRC baseline model. Without loss of generality, we take the BERT-based MRC model as an example to introduce our method, which is shown in the left part of Figure~\ref{fig:overall}.

Following~\citet{devlin2018bert}, given a data tuple $(q,d,s,e,t)$, we concatenate a ``[CLS]'' token, the tokenized question $q$ with length $n$, a  ``[SEP]'' token, the tokenized paragraph $d$ with length $m$, and a final ``[SEP]'' token. We feed the resulting sequence into the BERT model. The question $q$ and paragraph $d$ are first mapped into two sequence of embeddings:
\begin{align}
    \textbf{E}^q, \textbf{E}^d = \mathrm{BertEmbedding}(q,d),
\end{align}
where $\mathrm{BertEmbedding}(\cdot)$ denotes the BERT embedding layer which sums the corresponding token, segment, and position embeddings, $\textbf{E}^q \in \real{(n+2) \times h}$ and $\textbf{E}^d \in \real{m \times h}$ represent the question embedding and the paragraph embedding.

$\textbf{E}^q$ and $\textbf{E}^d$ are further fed into BERT layers which consist of multiple Transformer layers~\cite{vaswani2017attention} to obtain the final hidden representations $\{\textbf{C}, \textbf{T}^q_1,..., \textbf{T}^q_n,\textbf{T}_{[SEP]}, \textbf{T}^d_1,...,\textbf{T}^d_m\}$ as shown in Figure~\ref{fig:overall}. The representation vector $\textbf{C} \in \real{h}$ corresponding to the first input token ([CLS]) are fed into a binary classification layer to output the probability of whether the question is answerable:
\begin{align}
    P_a(\textrm{is-answerable})&=\mathrm{Sigmoid}(\textbf{C}\textbf{W}^T_c+\textbf{b}_c),
\end{align}
where $\textbf{W}_c \in \real{2 \times h}$ and $\textbf{b}_c \in \real{2}$ are trainable parameters.
The final hidden representations of paragraph $\{\textbf{T}^d_1,...,\textbf{T}^d_m\ \} \in \real{m \times h}$ are inputted into two classifier layer to output the probability of the start position and the end position of the answer span:
\begin{align}
    P_s(i=<\textrm{start}>) &= \mathrm{Sigmoid}(\textbf{T}^d_i\textbf{W}^T_s+b_s), \\
    P_e(i=<\textrm{end}>) &= \mathrm{Sigmoid}(\textbf{T}^d_i\textbf{W}^T_e+b_e),
\end{align}
where $\textbf{W}_s \in \real{1 \times h}$, $\textbf{W}_e \in \real{1 \times h}$, $b_s \in \real{1}$, and $b_e \in \real{1}$ are trainable parameters. 

For the data tuple $(q,d,s,e,t)$, the total loss of MRC model can be written as
\begin{align}
    \mathcal{L}_{\textrm{mrc}} &= \lambda \mathcal{L}_a(t) + \mathcal{L}_s(s) + \mathcal{L}_e(e), \\
    & =  -\lambda\log P_a(t) -\log P_s(s) -\log P_e(e), \notag
\end{align}
where $\lambda$ is a hyper-parameter.

\subsection{Transformer-based Autoencoder}\label{sec:ae}
As shown in the right part of Figure~\ref{fig:overall}, the original question $q$ is firstly mapped into question embedding $\textbf{E}^q$ with the BERT embedding layer. 
It should be noted that the Transformer encoder and the pre-trained MRC model share\footnote{The parameters of the Transformer encoder's embedding layer are copied from the pre-trained MRC, and are fixed during training.} the parameters of the embedding layer, which makes the question embedding of the two models in the same continuous embedding space.

We obtain the encoder hidden states $\textbf{H}_{enc} \in \real{(n+2) \times h}$ from the Transformer encoder. The objective of the Transformer autoencoder is to reconstruct the input question itself, which is optimized with cross-entropy~\cite{DaiL15}. 
A trivial solution of the autoencoder would be to simply copy tokens in the decoder side. 
To avoid this, we do not directly feed the whole $\textbf{H}_{enc}$ to the decoder, but use an RNN-GRU~\cite{cho2014learning} layer with sum pooling to obtain a latent vector $\textbf{z} \in \real{h}$. Then we feed $\textbf{z}$ to the decoder to reconstruct the question, which follows~\citet{wang2019controllable}. 
\begin{align}
    &\textbf{H}_{enc}=\mathrm{TransformerEncoder}(q), \\
    &\textbf{z} =\mathrm{Sum}(\mathrm{GRU}(\textbf{H}_{enc})), \\
    &\hat{q}=\mathrm{TransformerDecoder}(\textbf{z}).
\end{align}
We can train the autoencoder on the question data of $\mathcal{D}$ or pre-train it on other large-scale corpora, such as BookCorpus~\cite{zhu2015aligning} and English Wikipedia.

\begin{figure}[th]
    \centering
	\includegraphics[width = 3.1 in]{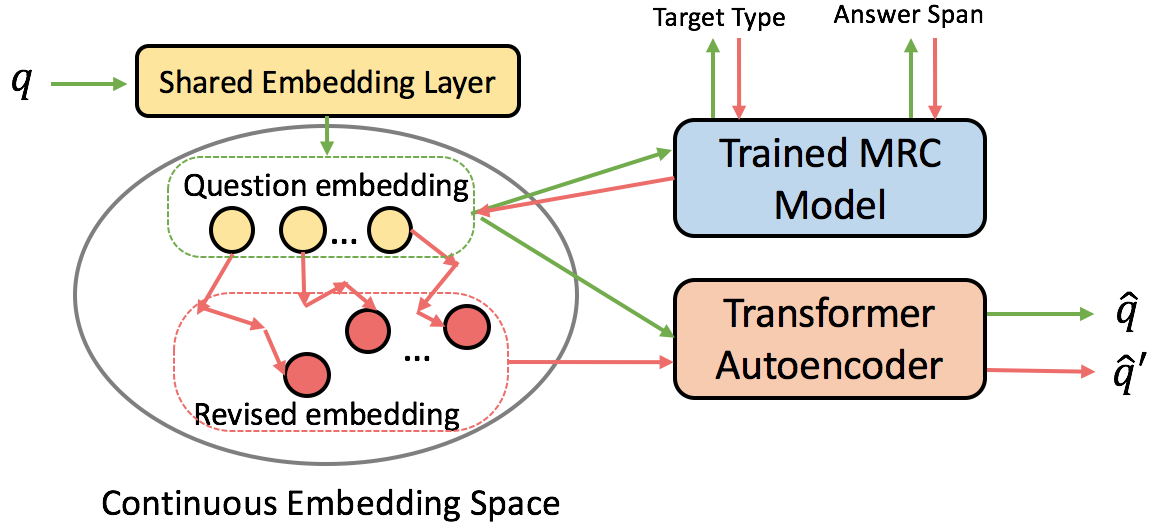}
	\caption{The question rewriting process of CRQDA.}\label{fig:inf}
\end{figure}

\begin{algorithm}[thb]
		\small
		\caption{Question Rewriting with Gradient-based Optimization.}
		\centering
		\label{alg}
		\begin{algorithmic}[1]
			\Require
			Data tuple $(q,d,s,e,t)$; Original question embedding $\textbf{E}^q$;
			pre-trained MRC model and Transformer autoencoder;
			A set of step size $S_{\eta}=\{\eta_i\}$;
			Step size decay coefficient $\beta_s$; 
			the target answerable or unanswerable label $t'$;
			Threshold $\beta_t, \beta_a, \beta_b$;
			\Ensure
			a set of new answerable and unanswerable question data tuples $\mathcal{D}'= \{(\hat{q}',d,s,e,t'),..,(\hat{q}',d,s,e,t)\}$;
			\State $\mathcal{D}' =\{\}$;
			\For{each $\eta \in S_{\eta}$ }
			\For{max-steps}
			\State revise $\textbf{E}^{q'}$ by Eq.~(\ref{eq:11}) or Eq.~(\ref{eq:10})
			\State $\hat{q}' = \textbf{TransformerAutoencoder}\left(\textbf{E}^{q'}\right)$
			\If{$P_a(t') > \beta_t$ and $ \mathcal{J}(q,\hat{q}') \in [\beta_a, \beta_b]$}
			\State add $(\hat{q}',d,s,e,t')$ to $\mathcal{D}'$;
			\EndIf
			\State $\eta = \beta_s \eta$;
			\EndFor
			\EndFor
			\\
		 \Return $\mathcal{D}'$;
		
		\end{algorithmic}
	\end{algorithm}
	
\subsection{Rewriting Question with Gradient-based Optimization}\label{sec:inf}
As mentioned above, the question embedding of the Transformer encoder and pre-trained MRC are in the same continuous embedding space, where we can revise the question embedding with the gradient guidance by MRC model. The revised question embedding $\textbf{E}^{q'}$ is fed into Transformer autoencoder to generate a new question data $\hat{q}'$. 

Figure~\ref{fig:inf} illustrates the process of question rewriting. 
Specifically, we take the process of rewriting an answerable question to a relevant unanswerable question as an example to present the process. 
Given an answerable question $q$, the goals of the rewriting are: (I) the revised question embedding should make the pre-trained MRC model predict the question from answerable to unanswerable with the paragraph $d$; (II) The modification size of $\textbf{E}^{q}$ should be adaptive to prevent the revision of $\textbf{E}^q$ from falling into local optimum; 
(III) The revised question $\hat{q}'$ should be similar to the original $q$, which helps to improve the robustness of the model. 

For goal-(I), we take the label $t'=0$, which denotes the label of question is unanswerable, to calculate the loss $\mathcal{L}_a(t')$ and the gradient of $\textbf{E}^q$ by the pre-trained MRC model (see the red line in Figure~\ref{fig:inf}). 
We iteratively revise $\textbf{E}^q$ with gradients from the pre-trained MRC model until the MRC model predicts the question is unanswerable with the revised $\textbf{E}^{q'}$ as its input, which means the $P_a(t'|\textbf{E}^{q'})$ is large than a threshold $\beta_t$. Note that here we use the gradient to only modify $\textbf{E}^q$, and all the model parameters during rewriting process are fixed. The process of each iteration can be written as:
\begin{align} \label{eq:10}
    \textbf{E}^{q'}=\textbf{E}^q- \eta (\nabla_{\textbf{E}^q} \mathcal{L}_a(t')), 
\end{align}
where $\eta$ is the step size. Similarly, we can revise the $\textbf{E}^q$ of a data tuple $(q,d,s,e,t)$ to generate a new answerable question whose answer is still the original answer span $(s,e)$ as follows:
\begin{align}\label{eq:11}
    \textbf{E}^{q'}=\textbf{E}^q- \eta \left(\nabla_{\textbf{E}^q} (\lambda\mathcal{L}_a(t)+\mathcal{L}_s(s) + \mathcal{L}_e(e))\right).
\end{align}
Rewriting the answerable question into another answerable question can be seen as a special constrained paraphrasing, which requires that the question after the paraphrasing is context-relevant answerable and its answer remains unchanged.

For goal-(II), we follow~\citep{wang2019controllable} to employ the dynamic-weight-initialization method to allocate a set of step-sizes $S_{\eta}=\{\eta_i\}$ as initial step-sizes.
For each initial step-size, we perform a pre-defined max-step revision with the step size value decay (corresponds to Algorithm \ref{alg} line 2-11) to find the target question embedding.
For goal-(III), we select the $\hat{q}'$ whose unigram word overlap rate with the original question $q$ is within a threshold range $[\beta_a, \beta_b]$. The unigram word overlap is computed by:
\begin{align}
    \mathcal{J}(q,\hat{q}') = \frac{\text{count}(w_q\cap w_{\hat{q}})}{\text{count}(w_q\cup w_{\hat{q}})},
\end{align}
here $w_q$ is the word in $q$ and $w_{\hat{q}}$ is the word in $\hat{q}'$. The whole question rewriting procedure is summarized in Algorithm~\ref{alg}.

\section{Experiments}
In this section, we describe the experimental details and results.
We first conduct the experiment on the SQuAD 2.0 dataset~\cite{rajpurkar2018know} to compare CRQDA with other strong baselines, which is reported in \S~\ref{sec:exp1}.
The ablation study and further analysis are provided in \S~\ref{sec:exp2}.
Then we evaluate our method on additional two tasks including question generation on SQuAD 1.1 dataset~\cite{rajpurkar2016squad} in \S~\ref{sec:exp3}, and question-answering language inference on QNLI dataset~\cite{wang2018glue} in \S~\ref{sec:exp5}.

\begin{table}[th]
\begin{center}
\small
  \begin{tabular}{lcccccl}
    \hline
    Methods & EM & F1 \\
    \hline
    BERT\textsubscript{large}~\cite{devlin2018bert} (original) & 78.7 & 81.9 \\
    \quad + EDA~\cite{wei2019eda} & 78.3 & 81.6 \\
    \quad + Back-Translation~\cite{yu2018qanet} & 77.9 & 81.2 \\
    \quad + Text-VAE~\cite{Liu2019ForcingTV} & 75.3 & 78.6 \\
    \quad + AE with Noise & 76.7 & 79.8 \\
    \quad + 3M synth~\cite{alberti2019synthetic} & 80.1 & 82.8 \\
    \quad + UNANSQ~\cite{zhu2019learning} & 80.0 & 83.0 \\
    \quad + CRQDA (ours) & \textbf{80.6} & \textbf{83.3} \\ 
    \hline
\end{tabular}
\end{center}
\caption{Comparison results on SQuAD 2.0.}
\label{tab:squad}
\end{table}

\subsection{SQuAD} \label{sec:exp1}
The extractive MRC benchmark SQuAD 2.0 dataset contains about 100,000 answerable questions and over 50,000 unanswerable questions. Each question is paired with a Wikipedia paragraph.
\paragraph{Implementation} Based on  \textit{RobertaForQuestionAnswering}\footnote{\url{https://github.com/huggingface/transformers}.} model of Huggingface~\cite{Wolf2019HuggingFacesTS}, we train a RoBERTa\textsubscript{large} model on SQuAD 2.0 as the pre-trained MRC model for CRQDA. The hyper-parameters are the same as the original paper~\cite{liu2019roberta}. 
For training the autoencoder, we copy the word embedding parameters of the pre-trained MRC model to autoencoder and fix them during training. Both of its encoder and decoder consist of 6-layer Transformers, where the inner dimension of feed-forward networks (FFN), hidden state size, and the number of attention head are set to 4096, 1024, and 16. 

The autoencoder trains on BookCorpus~\cite{zhu2015aligning} and English Wikipedia~\cite{devlin2018bert}. The sequence length, batch size, learning rate, and training steps are set to 64, 256, 5e-5 and 100,000. For each original answerable data, we use CRQDA to generate new unanswerable question data, resulting in about 220K data samples (including the original data samples). The hyper-parameter of $\beta_s$, $\beta_t$, $\beta_a$, $\beta_b$, and max-step are set to 0.9, 0.5, 0.5, 0.99, and 5, respectively.
\paragraph{Baselines} We compare our CRQDA against the following baselines: 
(1) \textbf{EDA}~\cite{wei2019eda}: it augments question data by performing synonym replacement, random insertion, random swap, or random deletion operation. We implement EDA with their source code\footnote{\url{https://github.com/jasonwei20/eda_nlp}.} to synthesize a new question data for each question of SQuAD 2.0;  
(2) \textbf{Back-Translation}~\cite{yu2018qanet,style_transfer_acl18}: it uses machine translation model to translate questions into French and back into English. We implement Back-Translation based on the source code\footnote{\url{https://github.com/shrimai/Style-Transfer-Through-Back-Translation}.} to generate a new question data for each original question; 
(3) \textbf{Text-VAE}~\cite{bowman2015generating,Liu2019ForcingTV}: it uses RNN-based VAE to generate a new question data for each question of SQuAD 2.0. The implementation is based on the source code\footnote{\url{https://github.com/dayihengliu/Mu-Forcing-VRAE}.};
(4) \textbf{AE with Noise}: it uses the same autoencoder of CRQDA for question data rewriting. 
The only difference is that the autoencoder cannot utilize the MRC gradient but only uses a noise (sampled from Gaussian distribution) to revise the question embedding.  
This experiment is designed to show necessity of the pre-trained MRC. 
(5) \textbf{3M synth}~\cite{alberti2019synthetic}: it employs round-trip consistency technique to synthesize 3M
questions on SQuAD 2.0;
(6) \textbf{UNANSQ}~\cite{zhu2019learning}: it employs a pair-to-sequence model to generate 69,090 unanswerable questions. 
Following previous methods~\cite{zhu2019learning, alberti2019synthetic}, we use each augmented dataset to fine-tune BERT\textsubscript{large} model, where the implementation is also based on Huggingface. 
\paragraph{Results} For SQuAD 2.0, Exact Match (EM) and F1 score are used as evaluation metrics. The results on SQuAD 2.0 development set are shown in Table~\ref{tab:squad}. 
The popular textual DA methods (including EDA, Back-Translation, Text-VAE, and AE with Noised), do not improve the performance of the MRC model. 
One possible reason might be that they introduce detrimental noise to the training process as they augment question data without considering the paragraphs and the associated answers. 
In sharp contrast, the QDA methods (including 3M synth, UNANSQ, and CRQDA) improve the model performance. 
Besides, our CRQDA outperforms all the strong baselines, which brings about 1.9 absolute EM score and 1.5 F1 score improvement based on BERT\textsubscript{large}. We provide some augmented data samples of each baseline in \textbf{Appendix A}.

\subsection{Ablation and Analysis}\label{sec:exp2}
Our ablation study and further analysis are designed for answering the following questions: \textbf{Q1}: How useful is the augmented data synthesized by our method if trained by other MRC models? \textbf{Q2}: How does the choice of the corpora for autoencoder training influence the performance? \textbf{Q3}: How do different CRQDA augmentation strategies influence the model performance?

\begin{table}[th]
\begin{center}
\small
  \begin{tabular}{llll}
    \hline
    Methods & EM & F1 \\
    \hline 
    BERT\textsubscript{base} & 73.7 & 76.3 \\
    \quad + CRQDA & \textbf{75.8} (+2.1) & \textbf{78.7} (+2.4) \\ 
    \hline
    BERT\textsubscript{large} & 78.7  & 81.9  \\
    \quad + CRQDA  & \textbf{80.6} (+1.9) & \textbf{83.3} (+1.4)\\ 
    \hline
    RoBERTa\textsubscript{base} & 78.6 & 81.6 \\
    \quad + CRQDA & \textbf{80.2} (+1.6) & \textbf{83.1} (+1.5) \\ 
    \hline
    RoBERTa\textsubscript{large} & 86.0 & 88.9 \\
    \quad + CRQDA  & \textbf{86.4} (+0.4) & \textbf{89.5} (+0.6) \\ 
    \hline
\end{tabular}
\end{center}
\caption{Results of different MRC models with CRQDA on SQuAD 2.0.}
\label{tab:squad_diff}
\end{table}

To answer the first question (\textbf{Q1}), we use the augmented SQuAD 2.0 dataset in \S~\ref{sec:exp1} to train different MRC models (BERT\textsubscript{base}, BERT\textsubscript{large}, RoBERTa\textsubscript{base}, and RoBERTa\textsubscript{large}). The hyper-parameters and implementation are based on Huggingface~\cite{Wolf2019HuggingFacesTS}. The results are presented in Table~\ref{tab:squad_diff}. We can see that CRQDA can improve the performance of each MRC model, yielding 2.4 absolute F1 improvement with BERT\textsubscript{base} model and 1.5 absolute F1 improvement with RoBERTa\textsubscript{base}. 
Besides, although we use a RoBERTa\textsubscript{large} model to guide the rewriting of question data, the augmented dataset can further improve its performance.

\begin{table}[th]
\begin{center}
\small
  \begin{tabular}{lcccccl}
    \hline
    Methods & EM & F1 & R-L & B4\\
    \hline 
    BERT\textsubscript{base} & 73.7 & 76.3 & - & -\\
    \quad + CRQDA (SQuAD 2)& 74.8 & 77.7 & 82.9 & 59.6\\ 
    \quad + CRQDA (2M ques)& 75.3 & 78.2 & 97.8 & 94.7 \\ 
    \quad + CRQDA (Wiki)& \textbf{75.8} & \textbf{78.7}  & 99.3 & 98.4\\ 
    \quad + CRQDA (Wiki+Mask)& 75.4 &78.4 & \textbf{99.7} & \textbf{99.4}\\ 
    \hline
\end{tabular}
\end{center}
\caption{Results of training autoencoder on different corpora. R-L is short for ROUGE-L, and B4 is short for BLEU-4.}
\label{tab:ae}
\end{table}

For the second question (\textbf{Q2}), we conduct experiments to use the following different corpora to train the autoencoder of CRQDA: (1) \textbf{SQuAD 2.0}: we use all the questions from the training set of SQuAD 2.0; (2) \textbf{2M questions}: we collect 2,072,133 questions from the training sets of several MRC and QA datasets, including SQuAD2.0, Natural Questions, NewsQA~\cite{trischler2016newsqa}, QuAC~\cite{choi2018quac}, TriviaQA~\cite{joshi2017triviaqa}, CoQA~\cite{reddy2019coqa}, HotpotQA~\cite{yang2018hotpotqa}, DuoRC~\cite{saha2018duorc}, and MS MARCO~\cite{bajaj2016ms}; (3) \textbf{Wiki}: We use the large-scale corpora English Wikipedia and BookCorpus~\cite{zhu2015aligning} to train autoencoder; (4) \textbf{Wiki+Mask}: We also train autoencoder on English Wikipedia and BookCorpus as \textbf{Wiki}. In addition, we randomly mask 15\% tokens of the encoder inputs with a special token, which is similar to the mask strategy used in~\cite{devlin2018bert,song2019mass}. 

We firstly measure the reconstruction performance of the autoencoders on the question data of SQuAD 2.0 development set. We use BLEU-4~\cite{papineni2002bleu} and ROUGE-L~\cite{lin2004rouge} metrics for evaluation.
Then we use these autoencoders for the CRQDA question rewriting with the same settings in \S~\ref{sec:exp1}. 
These augmented SQuAD 2.0 datasets are used to fine-tune BERT\textsubscript{base} model. 
We report the performance of fine-tuned BERT\textsubscript{base} model in Table~\ref{tab:ae}. 
It can be observed that with more training data, the reconstruction performance of autoencoder is better.
Also, the performance of fine-tuned BERT\textsubscript{base} model is better. 
When trained with Wiki and Wiki+Mask, the autoencoders can reconstruct almost all questions well. 
The reconstruction performance of model trained with Wiki+Mask performs the best. 
However, the fine-tuned BERT\textsubscript{base} model with autoencoder trained on Wiki performs better than that trained on Wiki+Mask. 
The reason might be that the autoencoder trained with denoising task will be insensitive to the word embedding revision of CRQDA. 
In other words, some revisions guided by the MRC gradients might be filtered out as noises by the autoencoder, which is trained with a denoising task.

\begin{table}[th]
\begin{center}
\small
  \begin{tabular}{lcccccl}
    \hline
    Methods & EM & F1 \\
    \hline 
    RoBERTa\textsubscript{large}~\cite{liu2019roberta} & 86.00 & 88.94 \\
    \quad + CRQDA (\textit{unans}, $\beta_a=0.7$) & 86.39 & 89.31\\ 
    \quad + CRQDA (\textit{unans}, $\beta_a=0.5$) & \textbf{86.43} & \textbf{89.50}\\ 
    \quad + CRQDA (\textit{unans}, $\beta_a=0.3$) & 86.26 & 89.35\\ 
    \hline
    \quad + CRQDA (\textit{ans}) & 86.22 & 89.30\\ 
    \quad + CRQDA (\textit{ans}+\textit{unans}) & 86.36 & 89.38\\ 
    
    \hline
\end{tabular}
\end{center}
\caption{Results of using differnt CRQDA augmented datasets for MRC training.}
\label{tab:ablation}
\end{table}

For the last question \textbf{Q3}, we use CRQDA for question data augmentation with different settings.  
For each answerable original question data sample from the training set of SQuAD 2.0, we use CRQDA to generate both answerable and unanswerable question examples. Then the augmented \textit{unanswerable} question data (\textit{unans}), the augmented \textit{answerable} question data (\textit{ans}), and all of them (\textit{ans} + \textit{unans}) are used to fine-tune RoBERTa\textsubscript{large} model. 
To further analyze the effect of $\beta_a$ (a larger $\beta_a$ value means that the generated questions are closer to the original question in the discrete space), we use different $\beta_a=0.3, 0.5, 0.7$ for question rewriting. The results are reported in Table~\ref{tab:ablation}. 
It can be observed that the MRC achieves the best performance when $\beta_a= 0.5$. Moreover, all of \textit{unans}, \textit{ans}, and \textit{ans} + \textit{unans} augmented datasets can further improve the performance. 
However, we find that the RoBERTa\textsubscript{large} model fine-tuned on \textit{ans} + \textit{unans} performs worse than fine-tuned on \textit{unans} only. 
The result is mixed in that using more augmented data is not always beneficial. 

\begin{table}[th] 
\small
\begin{center}
  \begin{tabular}{lcccccl} 
  \hline
    Method & B4 & MTR & R-L \\
\hline
 UniLM~\cite{dong2019unified} & 22.12    &25.06&   51.07\\
  ProphetNet~\cite{yan2020prophetnet} & 25.01  & 26.83   & 52.57\\ 
ProphetNet + CRQDA & \textbf{25.95}  & \textbf{27.40}   & \textbf{53.15}\\ \hline
 UniLM~\cite{dong2019unified} & 23.75    &25.61&   52.04\\
 ProphetNet~\cite{yan2020prophetnet} & 26.72  & 27.64   & 53.79 \\
 ProphetNet + CRQDA & \textbf{27.21}  & \textbf{27.81}  & \textbf{54.21} \\
\hline
\end{tabular}
\end{center}
\caption{Results on SQuAD 1.1 question generation. B4 is short for BLEU-4, MTR is short for METEOR, and R-L is short for ROUGE-L. The first block follows the data split in~\citet{du2017learning}, while the second block is the same as in \citet{zhao2018paragraph}.} \label{tab:qg}
\end{table}

\subsection{Question Generation} \label{sec:exp3}
Answer-aware question generation task~\cite{zhou2017neural} aims to generate a question for the given answer span with a paragraph.
We apply our CRQDA method to SQuAD 1.1~\cite{rajpurkar2016squad} question generation task to further evaluate CRQDA. The settings of CRQDA are the same as in \S~\ref{sec:exp1} and \S~\ref{sec:exp2}. The augmented answerable question dataset is used to fine-tune the ProphetNet~\cite{yan2020prophetnet} model which achieves the best performance on SQuAD 1.1 question generation task. The implementation is based on their source code\footnote{\url{https://github.com/microsoft/ProphetNet}.}. We also compare with the previous state-of-the-art model UniLM~\cite{dong2019unified}. Following~\citet{yan2020prophetnet}, we use BLEU-4, METEOR~\cite{banerjee2005meteor}, and ROUGE-L metrics for evaluation, and we split the SQuAD 1.1 dataset into training, development and test set. We also report the results on the another data split setting as in~\citet{yan2020prophetnet}, which reverses the development set and test set.
The results are shown in Table~\ref{tab:qg}. 
We can see that CRQDA improves ProphetNet on all three metrics and achieves a new state-of-the-art on this task. 

\begin{table}[t]
\begin{center}
\small
  \begin{tabular}{lcccccl}
    \hline
    Methods & Accuracy \\
    \hline
    BERT\textsubscript{large}~\cite{devlin2018bert} & 92.3 \\
    BERT\textsubscript{large} + CRQDA & \textbf{93.0}  \\ 
    \hline
\end{tabular}
\end{center}
\caption{Results on QNLI.}
\label{tab:qnli}
\end{table}

\subsection{QNLI} \label{sec:exp5}
Given a question and a context sentence, question-answering NLI asks the model to infer whether the context sentence contains the answer to the question. 
QNLI dataset~\cite{wang2018glue} contains 105K data samples. We apply CRQDA to QNLI dataset to generate new entailment and non-entailment data samples. 
Note that this task does not include the MRC model, but uses a text entailment classification model. 
Similarly, we train a BERT\textsubscript{large} model based on the code of \textit{BertForSequenceClassification} in Huggingface to replace the ``pre-trained MRC model'' of CRQDA to guide the question data rewriting. 
Following the settings in \S~\ref{sec:exp1}, we use CRQDA to synthesize about 42K new data samples as augmented data. Note that we only rewrite the question but keep the paired context sentence unchanged.
Then the augmented data and original dataset are used to fine-tine BERT\textsubscript{large} model. Table~\ref{tab:qnli} shows the results. 
CRQDA increases the accuracy of the BERT\textsubscript{large} model by 0.7\%, which also demonstrates the effectiveness of CRQDA.

\section{Conclusion}
In this work, we present a novel question data augmentation method, called CRQDA, for context-relevant answerable and unanswerable question generation. CRQDA treats the question data augmentation task as a constrained question rewriting problem. Under the guidance of a pre-trained MRC model, the original question is revised in a continuous embedding space with gradient-based optimization and then decoded back to the discrete space as a new question data sample. The experimental results demonstrate that CRQDA outperforms 
other strong baselines on SQuAD 2.0. The CRQDA augmented datasets can improve multiple reading comprehension models. Furthermore, CRQDA can be used to improve the model performance on question generation and question-answering language inference tasks, which achieves a new state-of-the-art on the SQuAD 1.1 question generation task.

\section*{Acknowledgment}
This work is supported in part by the National Natural Science Fund for Distinguished Young Scholar under Grant 61625204, and in part by the State Key Program of the National Science Foundation of China under Grant 61836006. 

\bibliography{emnlp2020}
\bibliographystyle{acl_natbib}

\appendix

\section{Augmented Dataset}
\begin{figure*}[th]
    \centering
	\includegraphics[width = 6.5 in]{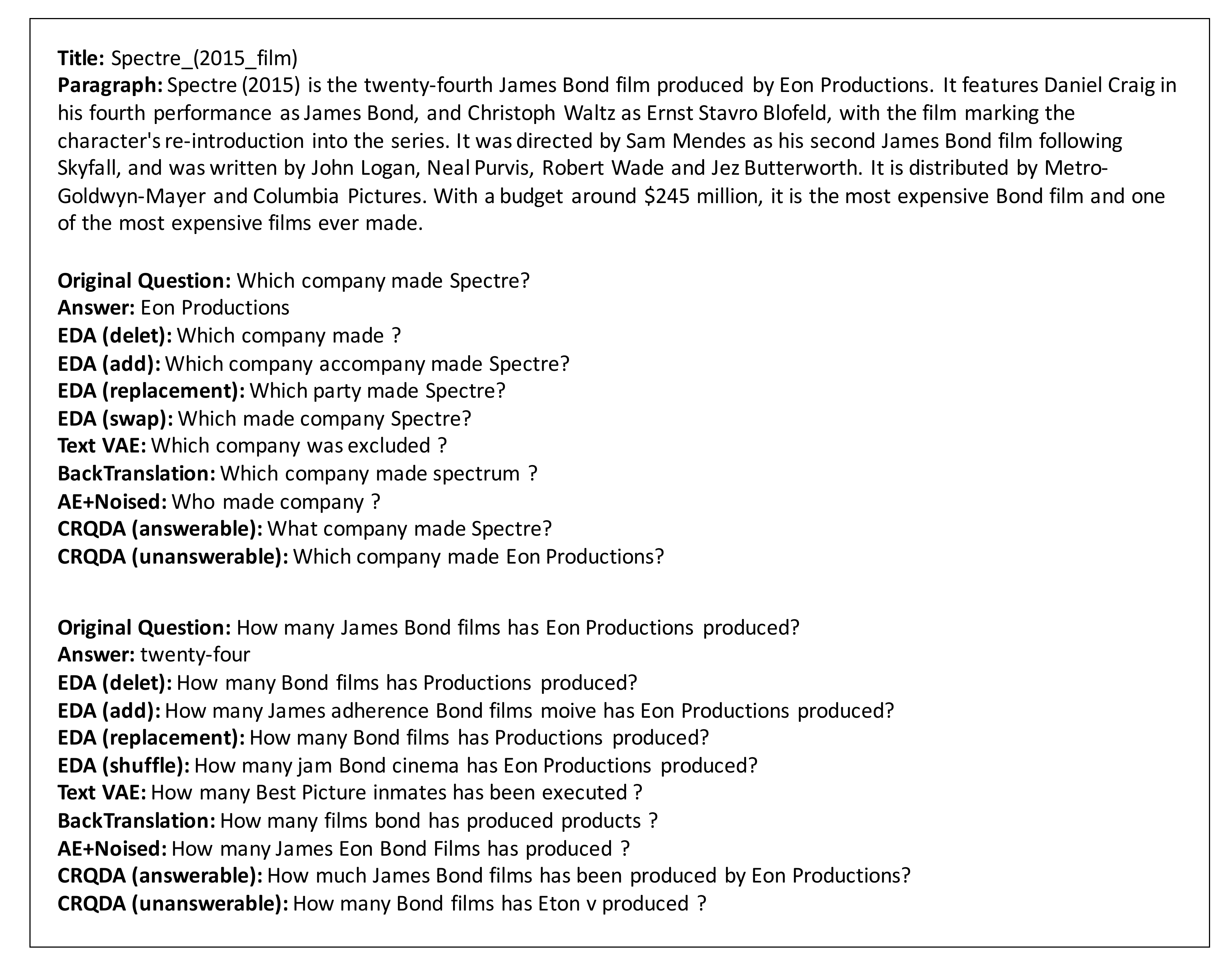}
	\caption{Augmented data samples on SQuAD 2.0. }\label{fig:1}
\end{figure*}

\begin{figure*}[th]
    \centering
	\includegraphics[width = 6.5 in]{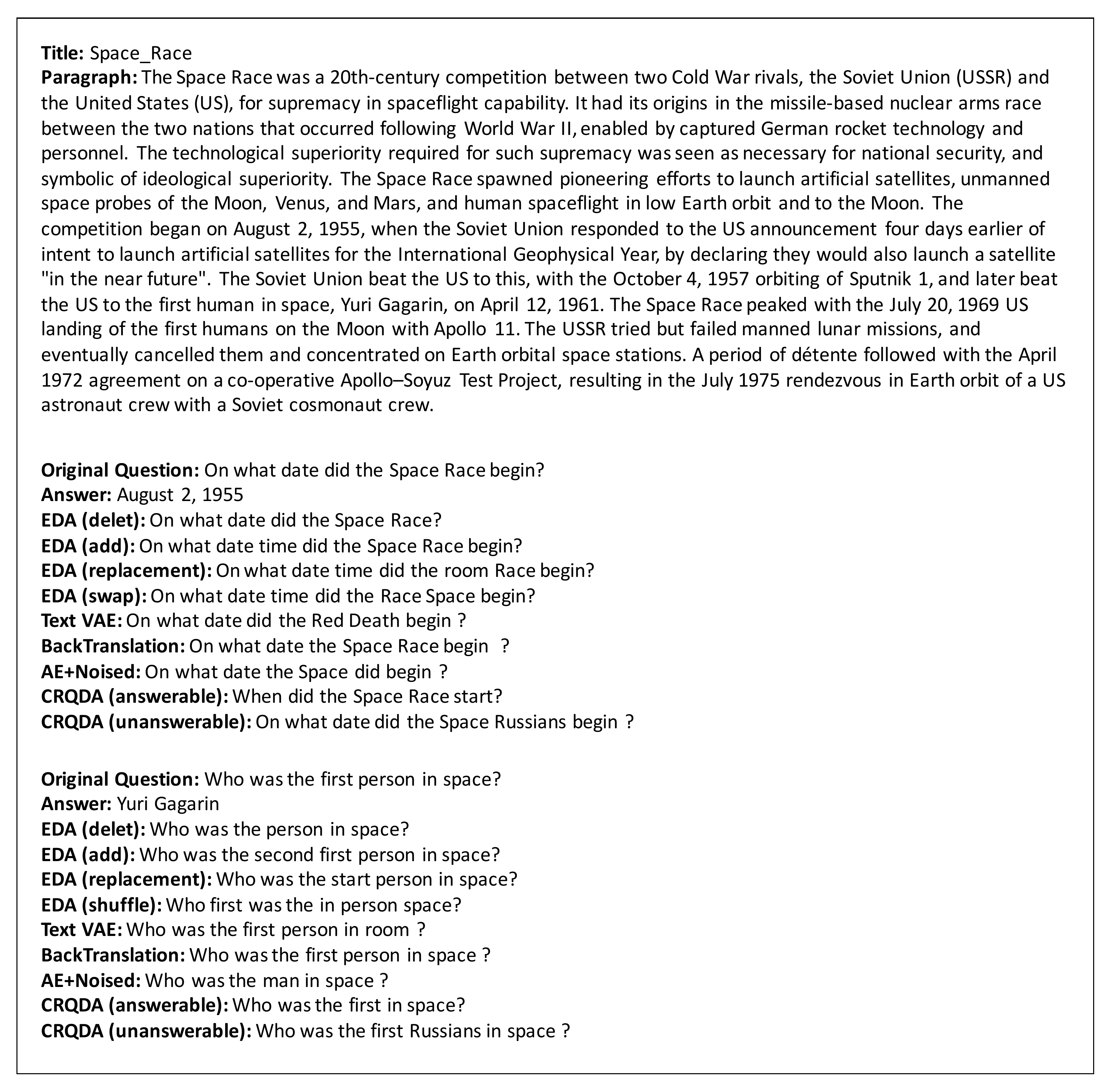}
	\caption{Augmented data samples on SQuAD 2.0. }\label{fig:2}
\end{figure*}

Figure~\ref{fig:1} and Figure~\ref{fig:2} provide some augmented data samples of each baseline on SQuAD 2.0. We can see that the baseline \textbf{EDA} tends to introduce noise which destroys the original sentence structure. The baselines of \textbf{Text VAE}, \textbf{BackTranslation} and \textbf{AE+Noised} often change some important words of the original question. This can cause the augmented question to miss the original key information and not to able to infer the original answer. 
In contrast, it can be observed that the generated answerable questions of \textbf{CRQDA} still maintain the key information for the original answer inference. 
Its generated unanswerable questions tend to introduce some context-relevant words to convert an original answerable question into an unanswerable one.

\end{document}